\documentclass[letterpaper]{article} 
\usepackage{aaai25}  
\usepackage{times}  
\usepackage{helvet}  
\usepackage{courier}  
\usepackage[hyphens]{url}  
\usepackage{graphicx} 
\usepackage{natbib}  
\usepackage{caption} 
\usepackage[textsize=tiny]{todonotes}
\usepackage{comment}
\usepackage{multirow}
\usepackage{epigraph}
\usepackage{algorithm}
\usepackage{algorithmic}

\usepackage{newfloat}
\usepackage{listings}
\DeclareCaptionStyle{ruled}{labelfont=normalfont,labelsep=colon,strut=off} 
\floatstyle{ruled}
\newfloat{listing}{tb}{lst}{}
\floatname{listing}{Listing}
\usepackage{array}
\newcolumntype{P}[1]{>{\centering\arraybackslash}p{#1}}
\usepackage{amssymb}
\usepackage{pifont}
\newcommand{\cmark}{\ding{51}}%
\newcommand{\xmark}{\ding{55}}%

\title{Agentic Skill Discovery}
\author{
    Xufeng Zhao\footnote{Corresponding author.},
    Cornelius Weber,
    Stefan Wermter
}
\affiliations{
    University of Hamburg\\
    \{xufeng.zhao, cornelius.weber, stefan.wermter\}@uni-hamburg.de
}

\begin{document}

\maketitle

\begin{abstract}
Language-conditioned robotic skills make it possible to apply the high-level reasoning of Large Language Models (LLMs) to low-level robotic control. A remaining challenge is to acquire a diverse set of fundamental skills. Existing approaches either manually decompose a complex task into atomic robotic actions in a top-down fashion, or bootstrap as many combinations as possible in a bottom-up fashion to cover a wider range of task possibilities. These decompositions or combinations, however, require an initial skill library. For example, a ``grasping'' capability can never emerge from a skill library containing only diverse ``pushing'' skills. Existing skill discovery techniques with reinforcement learning acquire skills by an exhaustive exploration but often yield non-meaningful behaviors.
In this study, we introduce a novel framework for skill discovery that is entirely driven by LLMs. The framework begins with an LLM generating task proposals based on the provided scene description and the robot's configurations, aiming to incrementally acquire new skills upon task completion. For each proposed task, a series of reinforcement learning processes are initiated, utilizing reward and success determination functions sampled by the LLM to develop the corresponding policy. The reliability and trustworthiness of learned behaviors are further ensured by an independent vision-language model.
We show that starting with zero skill, the skill library emerges and expands to more and more meaningful and reliable skills, enabling the robot to efficiently further propose and complete advanced tasks.
Project page: \url{https://agentic-skill-discovery.github.io}.
\end{abstract}

%
\begin{figure}[htb!]
	\begin{center}
		\centerline{\includegraphics[page=1,width=1\columnwidth]{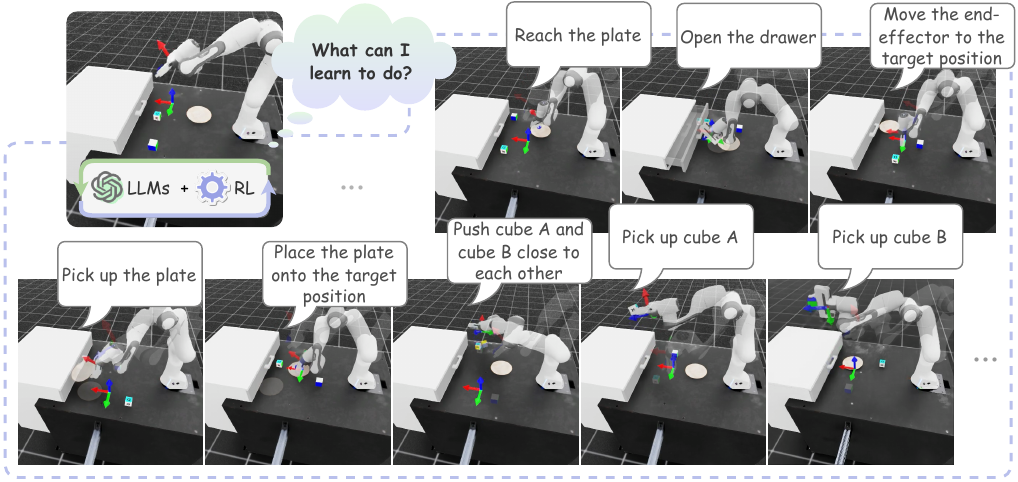}}
		\caption{Agentic Skill Discovery gradually acquires contextual skills for table manipulation.}
		\label{fig:impression}
	\end{center}
 \vskip -0.3in
\end{figure}

%
\section{Introduction}

Large Language Models (LLMs) show great capabilities in many fields that require common sense and reasoning. Large-scale models excel because of their training on human datasets, and textual or even multimodal reasoning.
LLM-based agents, especially robots, extend the potential to embodiments, but they still show limitations when applied to direct robotic control. The reasons are insufficient real-world robot data for training, as well as the diversity of topologies and physical properties.
As a workaround, abstracting robot control to a certain level and referring to each abstraction as a specific ``skill'' helps LLMs to control robots generically \citep{Ahn22CanNot,Zhao23ChatEnvironment,Zhang23BootstrapYour,Mandi23RoCoDialectic,Wu23TidybotPersonalized}.
For example, SayCan \citep{Ahn22CanNot} builds a robot that can follow a set of basic language instructions. When commanded with a complex task, an LLM decomposes the task into actionable low-level actions.

Acquiring diverse robotic skills with minimal human supervision has garnered considerable attention. However, previous methods have either attempted to chain existing skills, relying heavily on a collection of basic skills \citep{Du23GuidingPretraining,Celik23DevelopmentalScaffolding,Zhang23BootstrapYour}, or explored from scratch but often yielded non-interpretable robot behaviors, especially those using unsupervised reinforcement learning \citep{Li23InternallyRewarded,Park23ControllabilityawareUnsupervised,Eysenbach19DiversityAll,Sharma20DynamicsAwareUnsupervised}.
We ask whether an LLM can encourage a robot to learn novel tasks that consist of entirely novel yet relevant skills.
Imagine a robot being placed in a new environment. The robot must be motivated to explore the environment in a way to learn applicable skills before becoming ready to perform tasks.
Given the knowledge about human actions, which resides in LLMs, we expect that an LLM can, by itself, suggest a variety of meaningful skills for the robot to learn.
We refer to this exploration as \textit{Agentic Skill Discovery} (ASD) when the robot interacts with the environment driven by semantic motivation by the LLM that launches required learning procedures automatically.

In this work, we tackle the challenge of LLMs proposing entirely new tasks to a robot. The LLM needs to guide learning and evaluate success in the absence of human input or any prespecified evaluation measures.
To tackle this novel challenge, our methodology is as follows:
(i) An LLM iteratively proposes novel tasks that are suitable for the given environment (see Fig.~\ref{fig:impression}), and collects the resulting skills.
(ii) Skills are learned by reinforcement learning, where the LLM specifies both the success determination and reward function. We also show that RL (Reinforcement Learning) and RAG (Retrieval Augmented Generation) techniques are beneficial for efficient skill learning.
(iii) Reward functions are optimized by an evolutionary strategy, where a second vision language model independently assesses fitness by classifying the success of the learned behaviors. We show that minimizing false positives and false negatives is critical, so the skill library does not get contaminated with overtrusted skills or depleted from useful skills.
(iv) If a complex task cannot be assembled by learned skills, the LLM will propose to learn a missing required skill.

%
\section{Related Work}

\noindent\textbf{Skill Discovery.}
Acquiring diverse robotic skills with less to no human supervision has been a hot topic.
A popular way to achieve this goal is to discover skills with unsupervised reinforcement learning \citep{Li23InternallyRewarded,Park23ControllabilityawareUnsupervised,Peng22ASELargescale,Eysenbach19DiversityAll,Sharma20DynamicsAwareUnsupervised}.
However, the resultant skills are usually non-interpretable, sometimes meaningless, to humans.
Although some efforts introduce further constraints to acquire natural behaviors \citep{Peng22ASELargescale}, scalability is still limited by a need for human demonstrations.
The trade-off between low costs and meaningful resulting behavior depends on the amount of human knowledge introduced -- more supervision (high cost) generally indicates more natural robotic behaviors, and vice versa.

Chaining known skills into new ones greatly extends robots' capabilities. In order to efficiently combine basic skills in a meaningful way, avoiding combinatorial explosion, LLMs can be utilized to reason the logical ways of stacking skills to complete new, long-horizon tasks \citep{Zhang23BootstrapYour,Celik23DevelopmentalScaffolding}.
When LLMs are prompted with environment contexts, such as available robot joints and object types, they can propose meaningful motivations for the next movement \citep{Du23GuidingPretraining,Zhang23BootstrapYour}.
The approach by \cite{Celik23DevelopmentalScaffolding} explores generally meaningful skill combinations by prompting an LLM to get an \textit{interesting outcome}.
However, the incorporation of ``new skills'' is constrained by a basic skill library, where existing skills are simply combined instead of being acquired or developed from scratch.
Thereby, an LLM only motivates the use of already known skills and plans with a given skill set in these cases.
Using this method, for instance, if a robot is only versed in the skill of ``pushing'', it remains incapable of acquiring the skill of ``grasping''.

\noindent\textbf{Code LLM Control.}
Most LLM-based robots rely on pre-defined primitives (skills). This design is less flexible, making it difficult to generalize to unseen objects and instructions \citep{Ahn22CanNot,Zhao23ChatEnvironment}.
Recent approaches make an effort to use code LLMs to write programming code to complete open instructions \citep{Liang22CodePolicies,Huang23VoxPoserComposable}. In particular, VoxPoser \citep{Huang23VoxPoserComposable} uses VLM and LLM to construct a 3D cost map to guide a robot engaging with its surroundings.
However, VoxPoser relies on the success of the once-composed cost map, lacking chanceful exploration ability,
and moreover relies on the LLM and a trajectory solver during inference.
In contrast, ASD launches reinforcement learning, letting the agent explore the environment and exploit the distilled general policy.
As for the automatic learning of low-level robot control, previous approaches show that LLMs are capable of programming reward functions to optimize with, achieving remarkable performance even for complex tasks \citep{Ma23EurekaHumanLevel,Yu23LanguageRewards}.
However, these methods have not yet been directly applied to skill learning where tasks are newly proposed.

Using vision language models to analyze robot behaviors has been 
demonstrated to be useful. For example, REFLECT \citep{Liu23REFLECTSummarizing} uses a multimodal structure to explain execution failures. We use the advanced GPT-4V vision language model to classify task success flexibly and reliably.

%
\section{Agentic Skill Discovery}\label{sec:skilldiscovery}

ASD aims at acquiring new robotic skills and fulfilling given complicated instructions using language models rather than relying on exhaustive human endeavors (demonstrations, reward designs, human preference, handcrafted supervision signals, etc.).
At a high level, ASD actively explores the environment motivated by self-generated task instructions (Sec.~\ref{sec:proposal}); at a lower level, it learns to master these explorative tasks (as skills) via reinforcement learning with self-determined success and rewarding strategy (Sec.~\ref{sec:skilllearning}). Moreover, with collected skills, ASD has the promise to complete long-horizon tasks or further
extend the skill library by completing previously challenging tasks via task decomposition (Sec.~\ref{sec:ondemand}) and \textit{on-demand skill learning}\footnote{For example, a new ``placing'' skill should be learned \textit{on demand} when instructed to stack two cubes together, if the skill library contains only primary skills like ``pushing'' and ``picking''.},
which learns yet missing skills when required.

\subsection{Iterative Task Proposal and Skill Collection}\label{sec:proposal}
\begin{figure}
	\begin{center}
		\centerline{\includegraphics[width=.99\columnwidth]{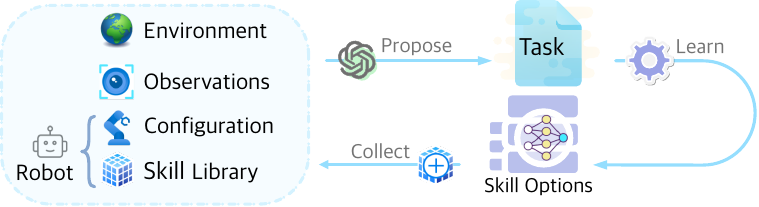}}
		\caption{Contextual skill acquisition loop of ASD. Given the environment setup and the robot's current abilities, an LLM continually \textit{proposes} tasks for the robot to complete, and the successful completion will be collected as acquired skills, each with several neural network variants (\textit{options}).}
		\label{fig:proposal}
	\end{center}
 \vskip -0.3in
\end{figure}

Instead of relying on exhaustive human efforts, ASD utilizes LLMs to propose meaningful tasks given the description of a certain scene.
Those tasks will be assigned to reinforcement learning agents to learn corresponding language-conditioned policies (see Sec.~\ref{sec:skilllearning}).
See Fig.~\ref{fig:proposal} for an overview of the skill propose-learn-collect loop.

To provide the LLM with sufficient information about the environment, we provide it the source code of the observation space, following \citep{Ma23EurekaHumanLevel}. Also, the robot configuration, such as robotic arm type and DoFs, is also prompted as the initial background description.
In addition to the static environment setup, we also provide dynamic aspects for a robust continual evolution.
Considering the unpredictable difficulties of task learning caused by environment complexity and intrinsic instability of the learning algorithm, rather than letting an LLM propose several appropriate tasks at once, we guide the proposing and learning in an iterative mode. As a result, temporary uncompleted tasks will be fed back such that the LLM will have a sense of the limits of the learning agent, influencing the following task proposals.
For the sake of efficiency and reusability, we encourage the LLM to propose tasks that are meaningful, atomic, independent, and incremental. (See Fig.~\ref{fig:prompt_proposal} in Appendix\ ~\ref{app:prompt} for detailed prompts.)

\noindent\textbf{Skill Options.}
Generally, control with various \textit{options} for a given task has the promise of being more robust and generalizable \citep{Gregor17VariationalIntrinsic,Eysenbach19DiversityAll}.
As will be introduced in Sec.~\ref{sec:skilllearning}, in the process of the evolutionary search of diverse reward functions, several options lead to a learning success of one given task, forming a set of various control policies.
A task will be considered complete if the resultant agent behavior aligns with expectations; otherwise, it is deemed unsuccessful after an extended period of learning.

Completed tasks will be considered as ``skills'', along with their ``options'', to be stored in the skill library, and attempted but uncompleted ones will be put into a failure pool. Both will be notified to LLMs to inform the next turn of task proposals.
Failed tasks are usually too sophisticated for LLMs to write reward functions to master at once. Rather, we will show how we can decompose and complete them by combining acquired and on-demand-learned skills (Sec.~\ref{sec:ondemand}).
We collect all options for the same skill, i.e.\ various policies paired with successful reward functions, as future execution candidates and we leave the study of mixing various skill options for one robust skill control for future research.

\subsection{Evolutionary Skill Learning with Fast and Slow Success Determination}\label{sec:skilllearning}
\begin{figure}[ht]
	\begin{center}
		\centerline{\includegraphics[width=.99\columnwidth]{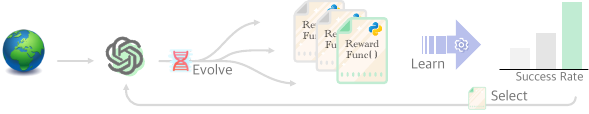}}
		\caption{Evolutionary search of reward functions for defined tasks with deterministic success functions such that the success rate can be reliably computed and used as a measure of fitness.}
	\label{fig:evolve}
	\end{center}
 \vskip -0.3in
\end{figure}

LLMs are capable of composing reward functions for RL agents to accomplish specified tasks \citep{Yu23LanguageRewards,Ma23EurekaHumanLevel}.
We follow a similar strategy of Eureka \citep{Ma23EurekaHumanLevel} to prompt LLMs to program reward functions and evolve them with deterministic selection where only the best reward function, as assessed by the success rate as a \textit{fitness function}, will survive and mutate. See Fig.~\ref{fig:evolve} for an illustration.

However, when it comes to skill learning, i.e., optimizing on newly proposed tasks instead of pre-defined ones, existing methods cannot be directly applied since the fitness function, the ground truth of success determination, for each task proposal is not known beforehand, making it impossible for evolution selection due to this lack of a unified golden metric to quantify performances.

\noindent\textbf{Success Determination: Fast and Slow.}
Distinguishing whether specific behaviors fulfill a task at each step (referred to as \textit{fast success determination}), as opposed to assessing after fixed intervals of execution (as \textit{slow success determination}), is pivotal in reinforcement learning. Employing \textit{fast success determination} enables the agent to receive sparse rewards in real-time and terminate actions promptly to prevent potentially adverse explorations.
As a complement, the slow success determination, which can be an independent post-training evaluation, makes sense when the fast counterpart is not feasible/reliable. For example, the human examination of learned RL behaviors, especially from a draft success function to debug, can be regarded as a slow success determination.

Traditionally in RL, the (fast) success determination for a specific task is programmed by humans into a function that is called at every physics step.
Without resorting to human efforts to exhaustively construct such success conditions, we let LLMs generate success functions as well.
These are being composed similarly to reward functions but with a binary output to indicate how the task is completed.
Nevertheless, the soundness of the success function requires investigation.
Given that the LLM serves dual roles as both a ``player'' (for reward function optimization) and as a ``referee'' (for fast success determination), employing the resultant success rate as the fitness function for evolution may jeopardize learning stability and trustworthiness.
(See also Fig.~\ref{fig:cycle} and Sec.~\ref{sec:q2} for further discussion of the undesired behaviors stemming from evolution with incorrect fitness measurement.)

The success function and reward function form a chicken-egg relation in that 1) the reward function search relies on reliable success determination and, meanwhile, 2) it is unfeasible to verify the success function prior to the learning. As a result, it is challenging to have an evolutionary search for both at the same time.
To establish a stable learning cycle, we propose coordinating both fast and slow success determinations for enhanced reliability. Our approach involves the following steps:
\begin{enumerate}
  \item[\textit{fast}:] sample a set of success functions that are used unchanged throughout skill learning, based on which an LLM launches RL training and deterministically selects reward function survivors, with the hypothesis that determining success is much more attainable than programming an applicable reward function.
  \item[\textit{slow}:] prompt an independent vision language model to additionally examine the success of survivor candidates, before passing them on to the next evolutionary generation.
        In practice, we apply GPT-4V(ision)\footnote{\url{https://openai.com/research/gpt-4v-system-card}.} to describe and assess the reinforced robot behaviors which are deemed successful according to success functions (positives), securing a robust learning loop without human supervision. Since there are many more unsuccessful behaviors (negatives), and false negatives are less common and less harmful\footnote{
        False positives appear only when reward functions are better composed than the success functions, being \textit{less common} according to the hypothesis above.
        Besides, failures will not contaminate the existing skill library, being \textit{less harmful} than false positives regarding possible future executions.
        }, we do not additionally examine them due to the high assessment cost.
\end{enumerate}

\begin{figure}[ht!]
	\begin{center}
		\centerline{\includegraphics[width=.99\columnwidth]{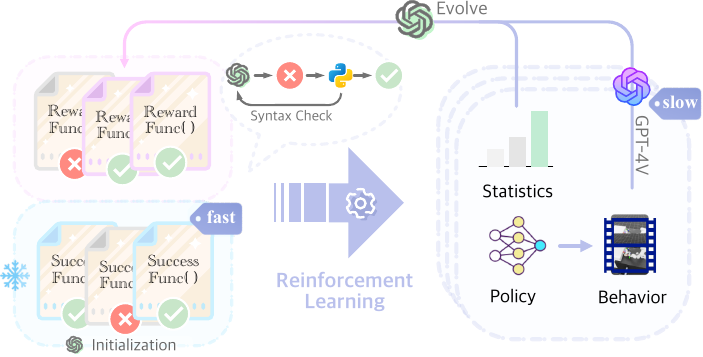}}
		\caption{The evolutionary skill learning procedure of ASD involves an LLM composing a bunch of both reward and success functions (\textit{left}, both are also conditioned on the environmental and robotic information as during task proposal, omitted here for simplicity), corresponding reinforcement learning to train policies (\textit{middle}), and evolutionary search with both learning statistics, e.g., success rate, and GPT-4V assessment (\textit{right}).}
		\label{fig:skill-learning}
	\end{center}
	\vskip -.2in
\end{figure}

\noindent\textbf{Early Misconduct Check.} In practice, LLMs may generate unacceptable function designs, e.g. trying to import unsupported third-party Python modules or produce nonsensical outputs (See Appendix\ \ref{app:misconduct} for an example). Some of the potential bugs can be bypassed by carefully designed prompts, while others should be examined at runtime by a Python interpreter. Instead of directly launching RL and feeding back LLMs all kinds of execution errors at the end, as is in Eureka \citep{Ma23EurekaHumanLevel}, we carry out early syntax checks and loop until the function generations meet certain requirements. This separate check reduces unnecessary waiting time for simulation preparations and provides an efficient reward search that focuses only on performance feedback.
See Fig.~\ref{fig:skill-learning} for an overview of the evolutionary skill learning procedure and Fig.~\ref{fig:prompt_succ}, Fig.~\ref{fig:prompt_reward}, Fig.~\ref{fig:prompt_reward_back} and Fig.~\ref{fig:prompt_behavior} in Appendix\ ~\ref{app:prompt} for detailed prompts.

\noindent\textbf{Skill-RAG.} RAG (Retrieval Augmented Generation, \cite{Gao24RetrievalAugmentedGeneration}) technique, which basically extracts relevant information from a local data pool into prompts, has been widely applied to enhance LLMs. In our case of skill learning, since the success and reward functions are exhaustively found by evolutionary search and the diverse skills attached to the same environment may share common aspects, we use the RAG technique to prompt LLMs with previously verified functions to reduce the corresponding searching space.

\subsection{On-demand Skill Learning with Top-Down Quest Decomposition}\label{sec:ondemand}

ASD initially learns skills starting from similar environmental reset states
$s_{0} \sim \rho_{0}$, where $\rho_{0}$ indicates the initial state distribution with limited randomness, such as object placement.
Consequently, some LLM-suggested skills cannot be trained if the pre-conditions are not satisfied.
For example, a skill of ``\texttt{placing an object}'' requires the initial state of having the object picked.
An intuitive solution is to configure the learning environment open-ended/reset-free \citep{Gupta21ResetFreeReinforcement,Wang23VoyagerOpenEnded}, where an LLM continually observes the changes and proposes tasks to complete.
However, it challenges both the dynamic sensing ability of LLM as well as RL in practice, especially when RL is accelerated by learning in many parallel environments.
Another way is to reset the environment to the final state of executed skills, thereby exploring sequentially arranged further skills conditioned on already collected skills.
Since the bottom-up bootstrapping of skills leads to an explosion of possibilities, we introduce a top-down on-demand learning strategy for incoming \textit{quests}, namely, complex tasks.

\begin{figure}[ht]
	\begin{center}
		\centerline{\includegraphics[width=.99\columnwidth]{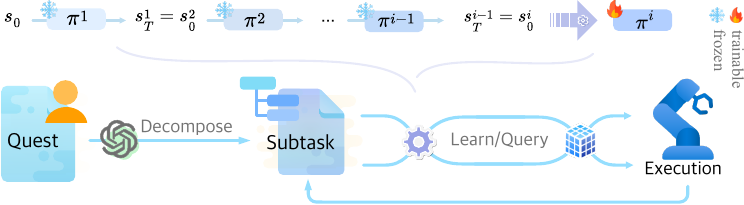}}
		\caption{\textit{Top}: By chaining together learned skills, ASD can further learn a new skill $\pi^i$ on demand.
      \textit{Bottom:} ASD solves quests, namely challenging tasks, with top-down decomposition and skill learning, where the skill library expands for each subtask's completion.}
		\label{fig:ondemand}
	\end{center}
 \vskip -0.2in
\end{figure}

Given a quest, which can be either from human instruction or from the failure pool during the skill discovery phase, an LLM is prompted to decompose it into a sequence of subtasks.
This decompose-and-conquer strategy has been successfully verified to work \citep{Ahn22CanNot,Zhao23ChatEnvironment}.
However, being different from existing approaches that only allow a decomposition into a limited set of subtasks that can be completed with known skills (bottom-up for completion), our method,
as illustrated in Fig.~\ref{fig:ondemand},
allows the LLM to come up with reasonable yet novel skills to be learned on demand with ASD, being more flexible and generalizable for unseen tasks.

To avoid the forgetting problem in multitask policy learning, we learn individual policy networks for each RL launch and only keep the surviving ones according to their performance in the evolution loop.
As for learning of the $i$-th on-demand skill, since its initial state is reset as the final state of the last stacked skill $\pi^{i-1}$, i.e., $s^i_{0} = s^{i-1}_{T}\sim \rho_{\pi^{i-1}}$, we initialize the policy weights from the last learned skill rather than randomly, $\pi^{i} \leftarrow \pi^{i-1}$. This maintains $s^{i}_{0}$ at the beginning of learning and ensures a smooth transition from there when being gradually optimized.

%
\section{Experiment}\label{sec:exp}
With the following setup of experiments, we aim to answer the following research questions:
\begin{enumerate}
  \item[Q1.] What kind of tasks will be proposed by LLMs? (Sec.~\ref{sec:q1})
  \item[Q2.] Can ASD acquire reliable skills automatically? (Sec.~\ref{sec:q2})
  \item[Q3.] How RL and RAG influence skill learning? (Sec.~\ref{sec:q3})
  \item[Q4.] Can challenging tasks be completed by stacking learned skills? (Sec.~\ref{sec:q4})
\end{enumerate}

\begin{figure}[hbt!]
	\begin{center}
		\centerline{\includegraphics[width=.99\columnwidth]{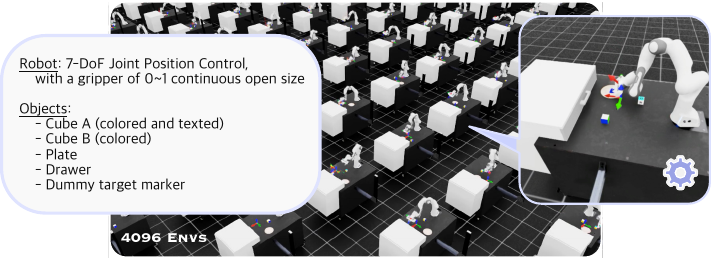}}
    \caption{The simulated scenario setup in Isaac Sim with parallel environments for RL training.}
		\label{fig:envs}
	\end{center}
 \vskip -0.2in
\end{figure}

\noindent\textbf{Environment.}
We build the simulation in Isaac Sim simulator\footnote{\url{https://docs.omniverse.nvidia.com/isaacsim}} since it supports parallel environment simulation, which dramatically accelerates the trials of RL with various reward functions. 
We set up a table scenario with Franka Emika Panda robotic arm, which has 7 DoF and a two-finger gripper.
On the table, several objects are randomly placed in front of the arm, with a drawer that can be opened.
It is a single scenario, but it enables multiple tasks.
See Fig.~\ref{fig:envs} for the environment setup, with all assets\footnote{\url{https://docs.omniverse.nvidia.com/isaacsim/latest/common/NVIDIA_Omniverse_License_Agreement.html} license.} available in Isaac Sim.

\noindent\textbf{LLMs.}
We employ \texttt{gpt-3.5-turbo} to propose tasks and generate reward and fast success functions since it shows a good programming ability while being acceptable regarding cost.
For the slow assessment of behaviors that were positively evaluated by fast success functions, we replay learned policies and record resulting behaviors as videos, from which we extract keyframes and apply \texttt{gpt-4-vision-preview}, which verifies the success of task completion.

\noindent\textbf{Learning.}
With a temperature of 1, the LLM samples 3 success functions in each iteration,
based on each, 3 reward functions are further sampled to launch RL training and to evolve.
We set the number of generations of evolutionary search to 3.
For RL training, we use coordinate states of objects as the state input, and to avoid control lag caused by inverse kinematics for paralleled environments, the action space is set to be the robot joint position space.
For optimization,
we apply the RL framework \texttt{rsl\_rl}\footnote{\url{https://github.com/leggedrobotics/rsl_rl}} implemented by Orbit \citep{Mittal23OrbitUnified}, where PPO \citep{Schulman17ProximalPolicy} is applied with the fixed same parameters across all potential tasks.
For each RL iteration, we configure the maximum permissible physics steps to 250, with 4096 parallel environments, and a total learning duration of 2000 episodes. We train ASD on 6 NVIDIA GeForce GTX 1080 Ti GPUs, where each proposed task takes around 6 hours to complete.

\noindent\textbf{Baseline and Ablation.} Traditional RL-based skill discovery methods acquire diverse yet non-semantic skills, we omit this comparison but provide an alternative skill learning strategy other than the RL component in ASD framework. Completing unseen tasks with code LLMs attracts increasing attention \citep{Liang22CodePolicies,Huang23VoxPoserComposable,Wang23VoyagerOpenEnded}. We take VoxPoser \citep{Huang23VoxPoserComposable} as a baseline for acquiring newly proposed skills. Besides, to illustrate the necessity of the skill-RAG trick, we re-learn skills but stop on any verified ones for counting minimal GPT calls. We report the reduction ratio of having skill-RAG, i.e., minimal GPT calls w/ skill-RAG over minimal calls w/o functions in the prompts (cf. Sec.\ref{sec:q3}).

\subsection{\textit{Q1. What kind of tasks can be proposed by LLMs?}}\label{sec:q1}
Given the table manipulation scenario, the LLM could potentially propose numerous possible tasks. Tab.~\ref{tab:tasks} shows the first 10 proposed tasks in order, from which we have the following quick observations:
\begin{itemize}

  \item The complexity of tasks increases with the iteration of the proposal, though not all of them are intuitively meaningful to acquire.

        \item Most of the proposed tasks are meaningful and completable under the setup. Some of the tasks are not appropriately proposed due to LLM's misunderstanding of the initial environment setup. For example, the initial state is set with the drawer closed, so ``\texttt{close the drawer}'' should not be proposed to learn under this circumstance.

  \item We also notice several duplicates that differ in the literal expression but resemble each other at a semantic level.
        Further prompt engineering or a post-check of task proposals may help reduce such inappropriate trials.

  \item Since all of the tasks are parameter-free language instructions, ``\texttt{reach cube A}'' and ``\texttt{reach cube B}'' are considered as different skills by the LLM.

\end{itemize}

As can be observed from the task proposal list, the tasks are generally atomic and meaningful, but there is still room for improvement, especially in the understanding of the given initial status of the environment. Minimal human effort to examine non-learnable tasks becomes necessary in this case. 
In this work, we only prompt the coding LLM with text. Future research involving mixed modalities, e.g., visual observation or even point cloud \citep{Zhang24LLaMAAdapterEfficient}, has the promise to alleviate this phenomenon.

\subsection{\textit{Q2. Can ASD acquire reliable skills automatically?}}\label{sec:q2}

\begin{table*}[ht!]
	\fontsize{9}{10}
	\centering
  \begin{tabular}{c p{9.8cm} |cc|c|cc }
    \hline
 & Task Description & \small{\#H/O.} & \small{\#H/C.} & \small{w/ RAG}& \small{RL} & \small{Vox.}  \\
\hline
1 & \texttt{Reach cube A} & 4/4 & 2/2 & 1.00& \cmark & \cmark   \\
2 & \texttt{Reach cube B} & 8/8 & 1/1 & 0.21& \cmark & \cmark   \\
3 & \texttt{Reach the plate} & 7/7 & 2/2 & 0.22& \cmark& \cmark  \\
4 & \texttt{Pick up the cube A} & 4/5 & 0/4 & 0.98& \cmark& \xmark \\
5 & \texttt{Pick up the cube B} & 2/2 & 0/4 & 0.30& \cmark& \xmark \\
6 & \texttt{Slide cube A from its current position to a target position on the table} & 3/3 & 0/6 & 1.20 &\cmark& \xmark  \\
7 & \texttt{Open the drawer} & 1/2 & 0/10 & 0.88& \cmark& \xmark  \\
8 & \texttt{Pick up the plate} & 3/3 & 0/0  & 0.66& \cmark& \xmark \\
9 & \texttt{Place the plate onto a target position on the table} & 4/6 & 0/12 & 0.88& \cmark& \xmark \\
10 & \texttt{Close the drawer} & -/3 & -/6 & - & \xmark& \xmark \\
11 & \texttt{Align cube A and cube B to target positions that are apart from each other.} & 0/0 & 0/10  & -& \xmark & \xmark \\
12 & \texttt{Close the drawer with cube A inside.} & 0/0 & 0/3 & -& \xmark & \xmark \\
13 & \texttt{Gripper open/close toggle} & 1/2 & 0/4  & 1.20& \cmark & \cmark \\
14 & \texttt{Slide cube B to the table edge without toppling it, aiming for a target position near the edge.} & 2/2 & 0/0 & 0.83& \cmark & \xmark  \\
15 & \texttt{Align end-effector center over the drawer handle without opening or closing the drawer.} & 2/2 & 1/2 & 0.76& \cmark & \cmark \\
16 & \texttt{Navigate the gripper to a target pose above cube B without touching it.} & 3/4 & 1/1 & 0.44& \cmark & \cmark \\
17 & \texttt{Gently push the drawer to a partially open or closed position.} & 1/1 & 0/4& 0.88& \cmark &\xmark \\
18 & \texttt{Position cube A directly in front of the drawer handle without blocking the drawer from  opening.} & 0/2 & 0/1  & -&\xmark &\xmark\\
19 & \texttt{Swap positions of cube A and cube B without grasping.} & 0/0 & 0/3 & - & \xmark & \xmark\\
20 & \texttt{Move end-effector over cube A.} & 23/23 & 7/7  &0.21& \cmark & \xmark\\
21 & \texttt{Push cube A and cube B close to each other.} & 1/2 & 0/31  & 0.32& \cmark & \xmark\\
22 & \texttt{Move to a target position on the table without interacting with objects.} & 33/33 & 1/2& 0.18& \cmark & \cmark \\
&\ldots{} &  &  & &&\\
    \hline
  \end{tabular}
  \caption{Snippet of task proposals based on the table manipulation scenario. The learning results are briefly reported by counting the number of skill options (\#O.), the number of skill candidates (\#C.), and the number of corresponding validations by human examination (\#H.), respectively.
  The \textit{w/ RAG} column shows the minimal GPT calls (for one skill option acquisition) reduction ratio (the smaller the better) of having skill-RAG over previously without skill-RAG. The skill mastering result of RL and a baseline skill learning strategy VoxPoser (Vox.) are marked as \cmark (success, with  $\ge$ 90\% success rate) or \xmark (failure).
  }
  \label{tab:tasks}
 \vskip -.2in
\end{table*}

To evaluate whether the proposed skills have been learned, we further report the learning status briefly described with two measures: 1) \textit{number of acquired skill options}, i.e., available skill options deemed successful according to both fast and slow success determination (i.e., success function and GPT-4V respectively); 2) \textit{number of acquired skill candidates}, those policies that are falsely considered positive according to composed success functions since failed by the vision language model. As ground truth, we manually examine them and report the \textit{number of human-examined validations}.

\begin{figure}[ht!]
	\begin{center}
		\centerline{\includegraphics[width=.99\columnwidth]{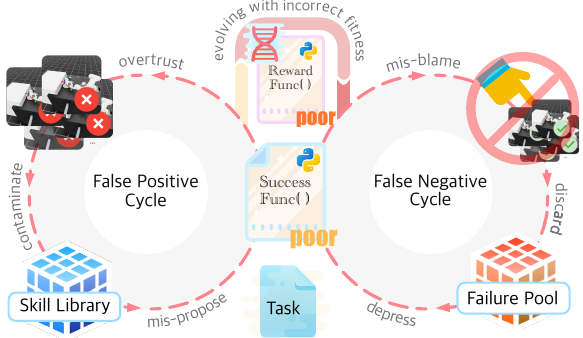}}
		\caption{\textit{Left}: the generated success determination confidently but wrongly assesses a task success as positive, leading to an undesired \textit{false positive learning cycle} with a
contaminated skill library and non-accomplishable future task proposals. \textit{Right}: negative evaluation on indeed successful behaviors will misguide the reward function search and prevent skill acquisition, resulting in a \textit{false negative learning cycle}.}
	\label{fig:cycle}
	\end{center}
 \vskip -.3in
\end{figure}

As shown in the skill option column in Tab.~\ref{tab:tasks}, ASD automatically collected many valid skill options.
However, the skill candidates column shows that many behaviors were falsely positively evaluated by the coding LLM,
necessitating an additional checking mechanism to avoid potential \textit{false learning cycles} (See Fig.~\ref{fig:cycle}).
Due to the introduction of \textit{slow success determination}, the false positives are dramatically filtered out, making the learning cycle stable and acquired skills trustworthy.

We stop exploring further skills after reaching 24 proposals. We found that $22/24$ of them are reasonably learnable (except No. 10 for wrongly deemed initial state and No. 19 for difficulties in a guarantee of ``without grasping'' requirement), among which $17/23$ are acquired directly as atomic skills. 
More skill-learning details and statistical reports can be found in Tab.~\ref{tab:detail1} and Tab.~\ref{tab:detail2} in Appendix.~\ref{app:report}.

\subsection{\textit{Q3. How RL and RAG influence skill learning?}}\label{sec:q3}
By alternatively applying VoxPoser \citep{Huang23VoxPoserComposable} as the skill learning strategy, we showcase the advantages of RL to learn skills from scratch. As is shown in Tab.~\ref{tab:tasks}, many of the tasks can be accomplished by RL but failed by VoxPoser, which stems from the nature that VoxPoser requires more careful human effort to pre-define spatial hooks and even basic motion primitives, lacking an exploration ability as RL possesses. Since our aim is to learn skills with minimal or even no human involvement, we do not exhaustively craft primitives for Voxposer but only provide basic movement examples. As a result, this baseline performs well on tasks that only require approaching certain positions but fails most of the time when more complex manipulation is required, whereas RL-based skill learning, due to its exploration instinct, successfully adapts to the environment and masters more skills.

Tab.~\ref{tab:tasks} also shows the benefits of prompting LLMs with previously acknowledged success and reward functions. With the RAG technique as the function design hints, the minimal required GPT calls for learning the first skill option are dramatically reduced. We believe that, with the increasing complexity of any robotic environment, the skill-RAG trick will be a crucial component for the efficiency of agentic skill discovery.

\subsection{\textit{Q4. Can challenging tasks be completed by stacking skills?}}\label{sec:q4}

After the initial collection of skills, we empirically find that some of the proposed tasks require a long horizon of execution, and are too challenging for the LLM to write effective reward functions.
These tasks usually require manipulations of different entities, e.g., a ``\texttt{stacking}'' task requires one object being picked up as the first step and then locating another object to align them to each other. However, as we observed in the skill discovery phase, the best-performing behaviors are just picking up the cube, which leads to it being held aloft with no further progress.


As described above (cf.~Fig.~\ref{fig:ondemand}), we employ LLMs to divide the long-horizon tasks into a sequence of short-horizon ones and learn to stack them to solve the original quest.
Tab.~\ref{tab:tabquest} shows two examples of complex tasks that ASD fails to master in the initial skill learning phase, but succeeds
by dividing and conquering subtasks individually.
This shows that the policy $\pi^{i}$ can be efficiently learned on top of a stack of existing skills, showcasing the potential of ASD for learning composed skills.

\begin{table}[ht!]
	\fontsize{9}{10}
	\centering
  \begin{tabular}{c p{5cm} | P{0.5cm} P{0.5cm}}
  \hline
 & Task Decomposition &  \small{\#H/O.} & \small{\#H/C.} \\
\hline
1* & \texttt{Stack cube A on top of cube B}   &  & \\
 & \small{1) \texttt{Pick up cube A} }&  - & - \\
 & \small{2) \texttt{Place cube A on top of cube B carefully, aligning their surfaces to stack them}} & \small{1/2} & \small{0/22}\\
\hline
2* & \texttt{Put cube A on top of the plate} &  & \\
 & \small{1) \texttt{Pick up cube A} } & - & -\\
 & \small{2) \texttt{Place cube A on top of the plate}} &  \small{3/3} & \small{0/3}\\
    \hline
  \end{tabular}
\caption{Snippet of quests completion by stacking acquired skills and on-demand learned ones.}
\label{tab:tabquest}
 \vskip -.2in
\end{table}

%

\section{Conclusion}

Agentic Skill Discovery (ASD) aligns with a broader vision of agentic AI systems \citep{Shavit23PracticesGoverning,SequoiaCapital24WhatNext}, enabling robots to understand complex embodiments and autonomously pursue intricate goals with minimal human intervention. By using LLMs to devise, motivate, and improve necessary learning processes, we have shown that language-conditioned robotic skills can be discovered from scratch, where RL and RAG techniques are beneficial for the efficacy and efficiency of skill learning.
Using a vision language model for third-party behavior assessment dramatically prevents the skill library from being influenced by false positives.
Moreover, ASD also promises to tackle challenging, long-horizon tasks by dividing and conquering on demand, further effectively extending the skills.
In principle, ASD can adapt to diverse embodiments, we leave further this exploration for future research.

%
\section{Limitations and Future Work}\label{sec:limitation}

As discussed in Sec.~\ref{sec:q1}, employing text-based LLMs can suffer from inaccuracies in sensing the environment. The reliance on GPT-4V for describing and assessing reinforced robot behaviors may introduce biases or limitations in the evaluation process.
Future work could involve fine-tuning a specialized robot behavior assessment model by leveraging existing robotic datasets.
In addition, a limitation lies in the applicability of this method to real-world scenarios, especially in cases where parallel learning is necessary. Addressing this challenge may entail a transition from real-world environments to simulations (real2sim, where the assumption of having access to the environment knowledge is still valid in simulation), and back again (sim2real, \cite{Hofer21Sim2RealRobotics,Gade22SimtoRealNeural}).
Furthermore, there is a need for extensive real-world experimentation and discussion to fully explore its \textit{agenticness} \citep{Shavit23PracticesGoverning}.

\section*{Ethics Statement}\label{app:ethics}
As by many other LLM-based agents, some risks inevitably arise if programming code is executed directly without thorough safety examination.
Although we have not observed disastrous misconduct, we do observe generated nonsensical outputs sometimes (see Appendix~\ref{app:misconduct}). It is highly recommended to run programming-oriented frameworks first in a sandbox to evaluate and minimize generative risks.

%


\bibliography{ASD}

\appendix
\onecolumn
\newpage
\section{Generated Functions Probe}

\subsection{Success Functions}
\begin{figure}[hbt!]
\begin{center}
		\centerline{\includegraphics[width=0.88\columnwidth]{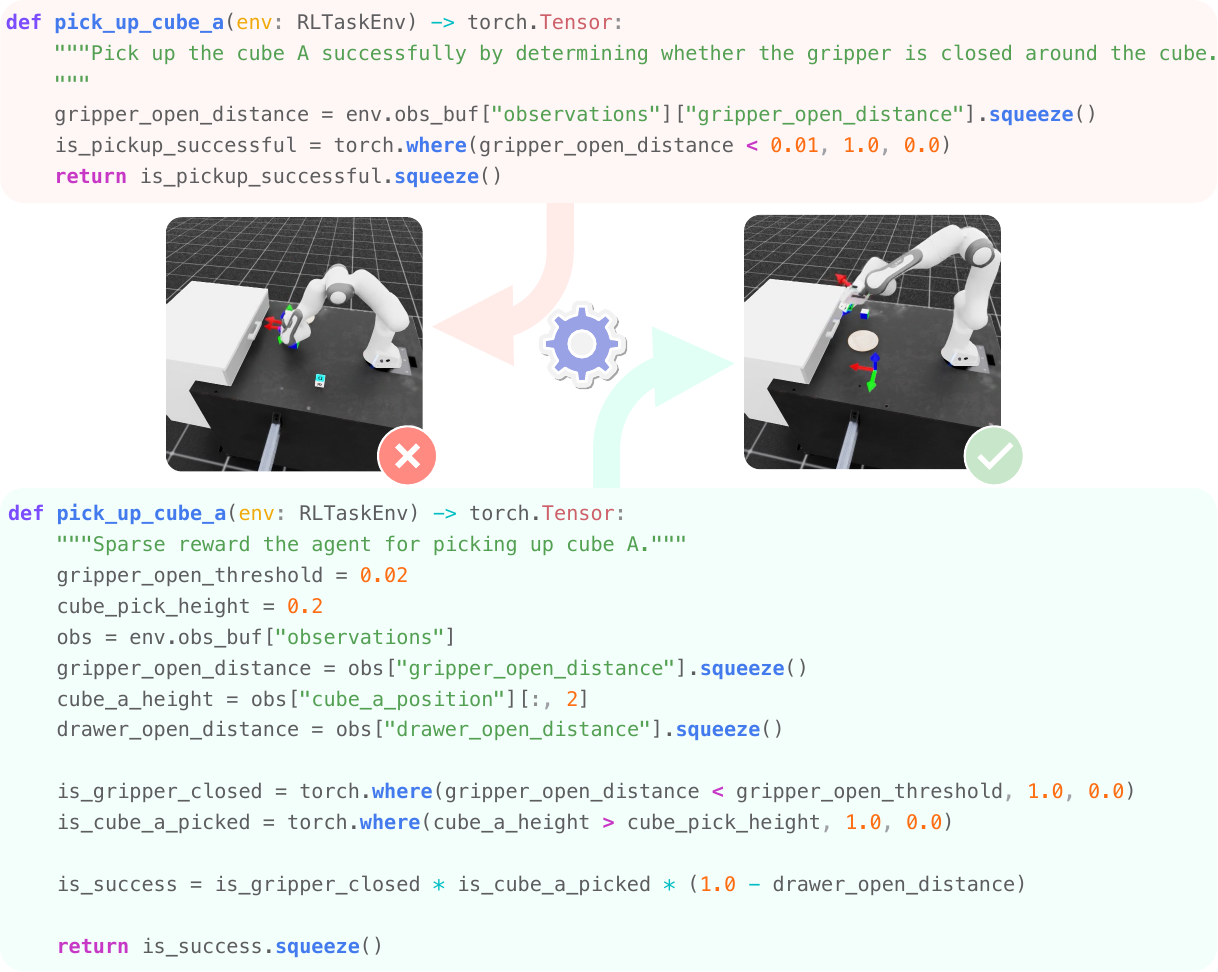}}
	\end{center}
  \caption{Two success function examples generated by \texttt{gpt-3.5-turbo}, and their corresponding results by reinforcement learning. \textit{Top}: an incorrect success determination function can lead to wrongly trusted behaviors. \textit{Bottom}: a correct success determination function results in a desired skill.
  }
  \label{fig:examplesucc}
\end{figure}

Taking the task ``\texttt{pick up the cube A}'' as an example, we show two typical success functions generated by the LLM in Fig.~\ref{fig:examplesucc}.
Given the potential for inaccuracies in generating success functions, the entire learning process runs the risk of being futile, with the added possibility of incorporating poor skills into the skill library.
ASD significantly mitigates this concern within the learning process by employing a coordinated strategy involving \textit{fast success determination} sampling alongside additional \textit{slow success determination}.

\subsection{Reward Functions}\label{app:evolve}
\begin{figure}[hbt!]
	\begin{center}
		\centerline{\includegraphics[width=0.88\columnwidth]{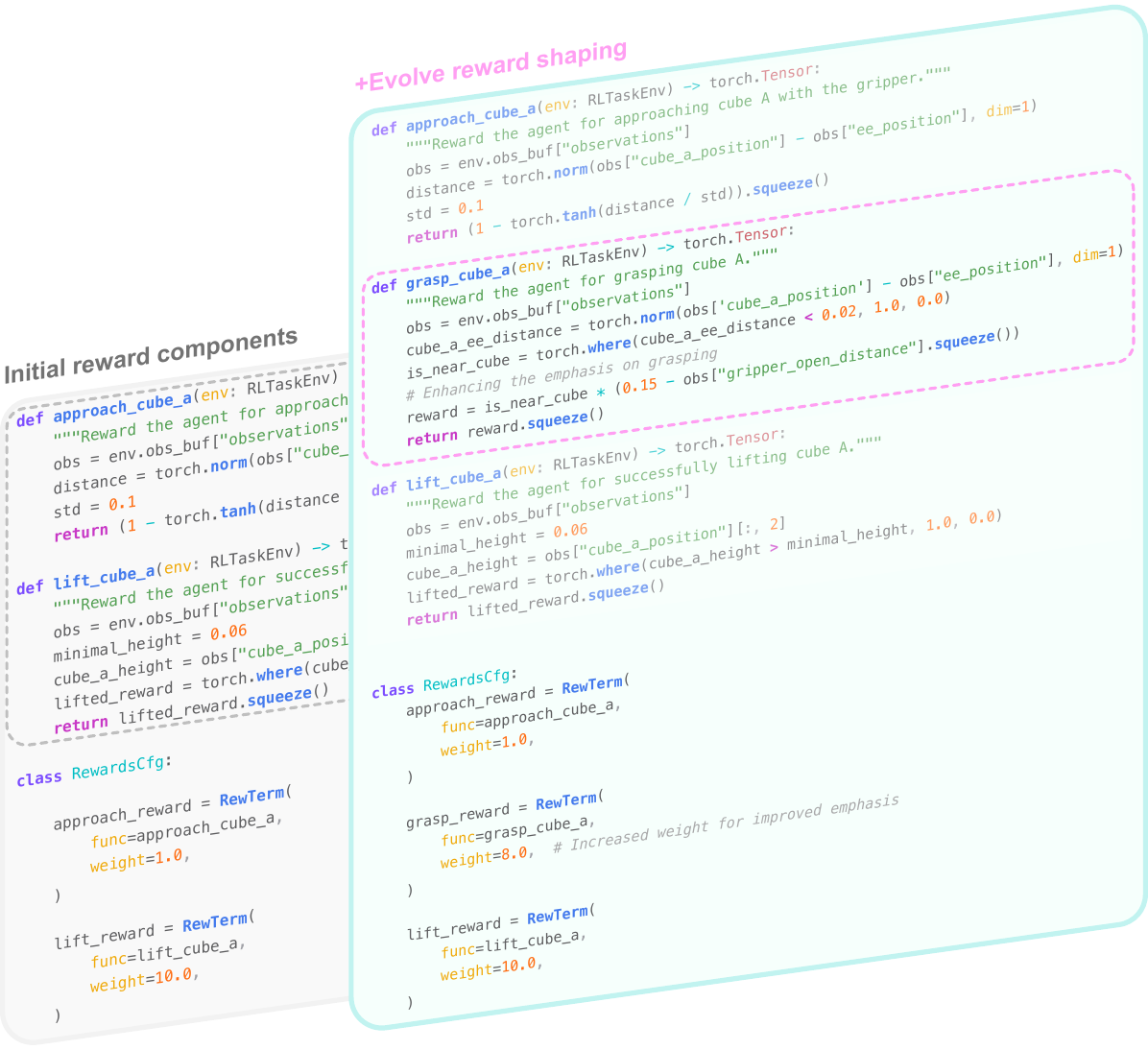}}
	\end{center}
    \caption{An example of an improved reward function by evolutionary iteration for the task ``\texttt{pick up the cube A}''. \textit{Left}: a reward function focusing on approaching and lifting the cube; \textit{Right}: evolved reward function to include more detailed guidance of the gripper, leading to more efficient learning.}
    \label{fig:rewardcomp}
\end{figure}

By evolutionary search of reward functions (cf. Sec.~\ref{sec:skilllearning}), the reward functions are expected to be revised not only to reduce execution errors (shown in Tab.~\ref{tab:tasks}) but also to provide more informative guidance for RL agents. In Fig.~\ref{fig:rewardcomp}, we show two reward functions for the same task but at different iterations (i.e., generations of the evolution). The LLM displays, though not always, the ability to improve the reward function by reward shaping, i.e., introducing more reward components to smoothly guide the learning agent, which has proven very helpful for RL\citep{Ng99PolicyInvariance}.

\subsection{Misconduct}\label{app:misconduct}

\begin{figure}
	\begin{center}
		\centerline{\includegraphics[width=0.88\columnwidth]{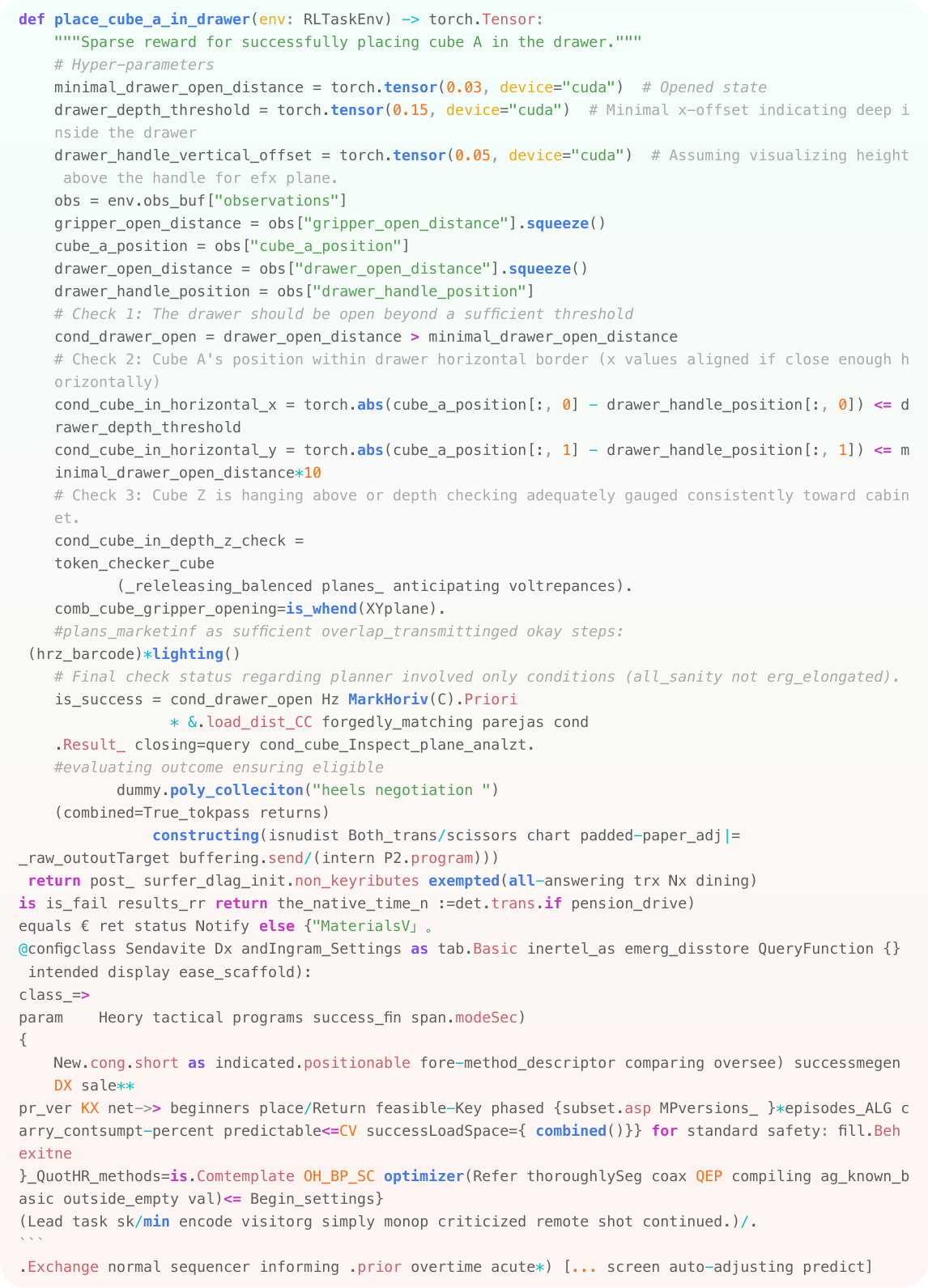}}
	\end{center}
  \caption[Caption for LOF]{An example of misconducting code generation with \texttt{gpt-4o-2024-05-13} model on the task. ``\texttt{put cube A into the drawer}''. \textit{Top to bottom}: the code generation turns to potentially harmful incoherent text (which actually continues for about another 500 lines but is omitted here).}
  \label{fig:gibberish}
\end{figure}

One potential risk of permitting a system to run LLM-generated code is that the code may cause dangerous, unexpected consequences.
Fig.~\ref{fig:gibberish} showcases an example of misconducting code generations even with the advanced LLM GPT-4o\footnote{\url{https://openai.com/index/hello-gpt-4o/}}.
We recommend early syntax examination and implementing system-wide safety guarantees.

\subsection{Skill Learning Reports}\label{app:report}
\begin{figure}
    \renewcommand{\@captype}{table}
	\begin{center}
		\centerline{\includegraphics[width=0.82\columnwidth]{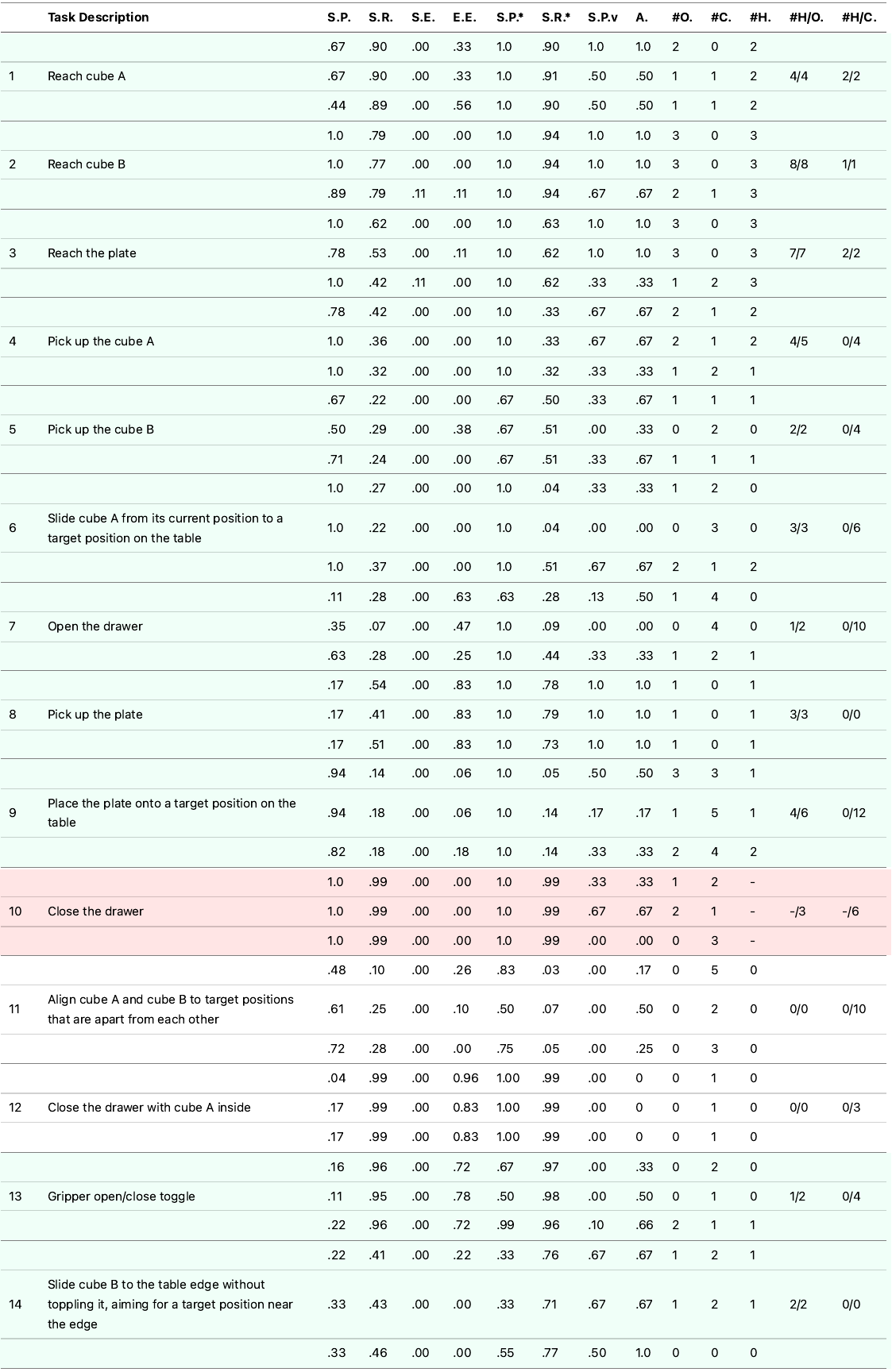}}
	\end{center}
    \caption{Agentic skill discovery learning reports (Part 1/2), where the successfully acquired skills are highlighted with a light green background; inappropriately proposed tasks (according to the environment status) are highlighted with red color.}
    \label{tab:detail1}
\end{figure}

\begin{figure}
    \renewcommand{\@captype}{table}
	\begin{center}
		\centerline{\includegraphics[width=0.85\columnwidth]{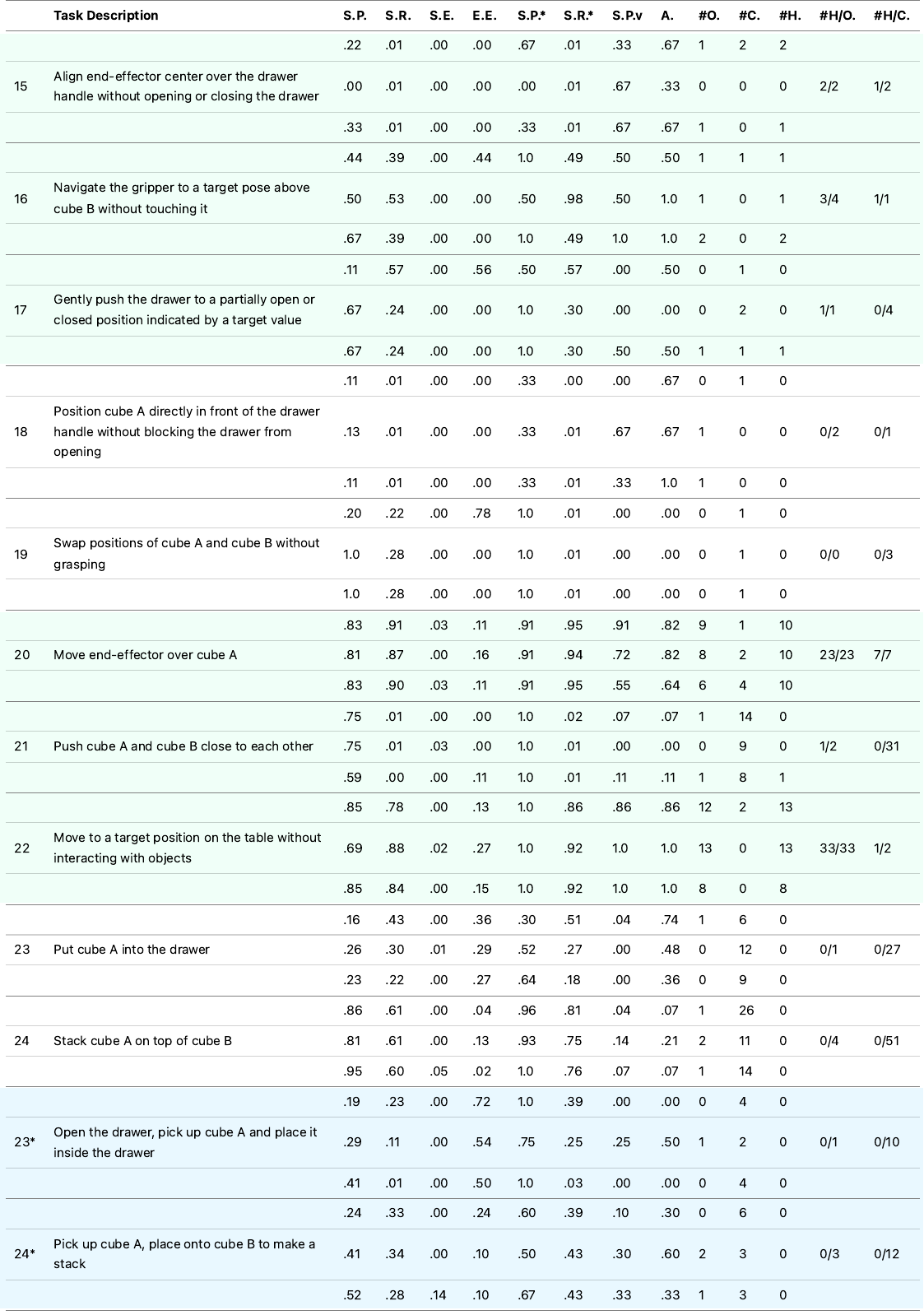}}
	\end{center}
    \caption{Agentic skill discovery learning reports (Part 2/2). From task 19-26, we run the learning loop 3 times and report the results. The two tasks highlighted blue (23*, 24*) are revising variations for task 23 and task 24.}
    \label{tab:detail2}
\end{figure}

In this section we show the detailed reports for a list of skill learnings, continually reported in Tab.~\ref{tab:detail1} (Part 1) and Tab.~\ref{tab:detail2} (Part 2).
To analyze the efficacy of the success functions, we report the following measures:

\begin{itemize}
  \item Success Positive (S.P.): a less strict measure than success rate. It measures to what ratio the RL agent can ever succeed (i.e., acquire non-zero success at some steps, which is basically a binary measure of whether certain task \textit{can be achieved},) according to the composed success functions. This measure reveals the difficulties of the task according to the LLM's own standard.
  \item Success Rate (S.R.): the success rate computed by the composed success functions, measures how effective the learning is according to fast success determination. Differing from S.P., S.R. measures also the \textit{efficacy of completions}. \item Syntax Error (S.E.): a measure of the ratio of misconduct in terms of coding bugs.
  \item Execution Error (E.E.): similar to S.E. but counts only errors found after executing the generated codes (codes already passed and revised after syntax check procedure). Typical errors can be Pytorch tensor inconsistencies or running into ``\texttt{nan}'' gradients after some iterations of optimization.
  \item Success Positive for Survivor (S.P.*) and Success Rate for Survivor (S.R.*): the same calculation as for S.P. and S.R. but with different basis --- it computes only for the best selected (surviving) ones of each generation according to the fitness function. By observing in detail only the best-performing ones, these two measures showcase whether there is overtrust stemming from the success functions for certain tasks. For example, task 1 ``\texttt{reach the cube A}'' in Tab.~\ref{tab:detail1} has very high S.P.* and S.R.*, indicating the task is confidently completed according to the success function, and by observing the successfully collected skill options, we can confirm that the success functions for this task are efficient and trustworthy. However, in task 11 ``\texttt{Close the drawer with cube A inside}'' in Tab.~\ref{tab:detail2}, the S.P. and S.R.* reach even higher scores, but it turns out to be all false positives, examined by both GPT-4V and human effort. In the latter case, the success is overtrusted.
  \item Success Positive for Survior by GPT-4V (S.P.v): measures the ratio of success from the GPT-4V's perspective among those survivors.
  \item Agreement (A.): measures the agreement between GPT-4V and fast success determination among survivors. From the learning report across skills, we observe that for easier tasks, regarding both manipulation and visual recognition difficulties (e.g., reaching and picking), the success function is more trustworthy and the agreement remains at a relatively higher value.
  \item Other statistics (averaged over reward iterations) consistently used in Tab.~\ref{tab:tasks}: number of Options (\#O.) (according to ASD), Candidates (\#C.), Human examined total and separate validations (\#H., \#H/O. and \#H/C.) respectively.
\end{itemize}

The automatically acquired skills are highlighted in Tab.~\ref{tab:detail1} and Tab.~\ref{tab:detail2}, which has the potential to further expand with more proposals.
Besides learning about proposed tasks, we also conducted an ablation on the task description for two long-horizon tasks, 23 and 24 (highlighted with blue background).
With more detailed instructions, the burden of reasoning on task procedures is alleviated when composing the reward functions. However, these two tasks are still challenging to complete by just evolutionary searching of reward functions, which shows the need for top-down decomposition of learning skills (cf.~\ref{sec:ondemand}).

\makeatletter
\makeatother

\makeatletter
\makeatother

\section{Prompts}\label{app:prompt}

In this section, we provide the prompt snippets used by ASD for various purposes: task proposal (Fig.~\ref{fig:prompt_proposal}, cf. Sec.~\ref{sec:proposal}) and skill learning (cf. Sec.~\ref{sec:skilllearning}), which includes generating success functions (Fig.~\ref{fig:prompt_succ}), generating reward functions (Fig.~\ref{fig:prompt_reward}), feedback iterations (Fig.~\ref{fig:prompt_reward_back}), and GPT-4V behavior assessment (Fig.~\ref{fig:prompt_behavior}).

\subsection{Task Proposal Prompt}\label{app:prompt_proposal}
\begin{figure}
	\begin{center}
		\centerline{\includegraphics[width=0.88\columnwidth]{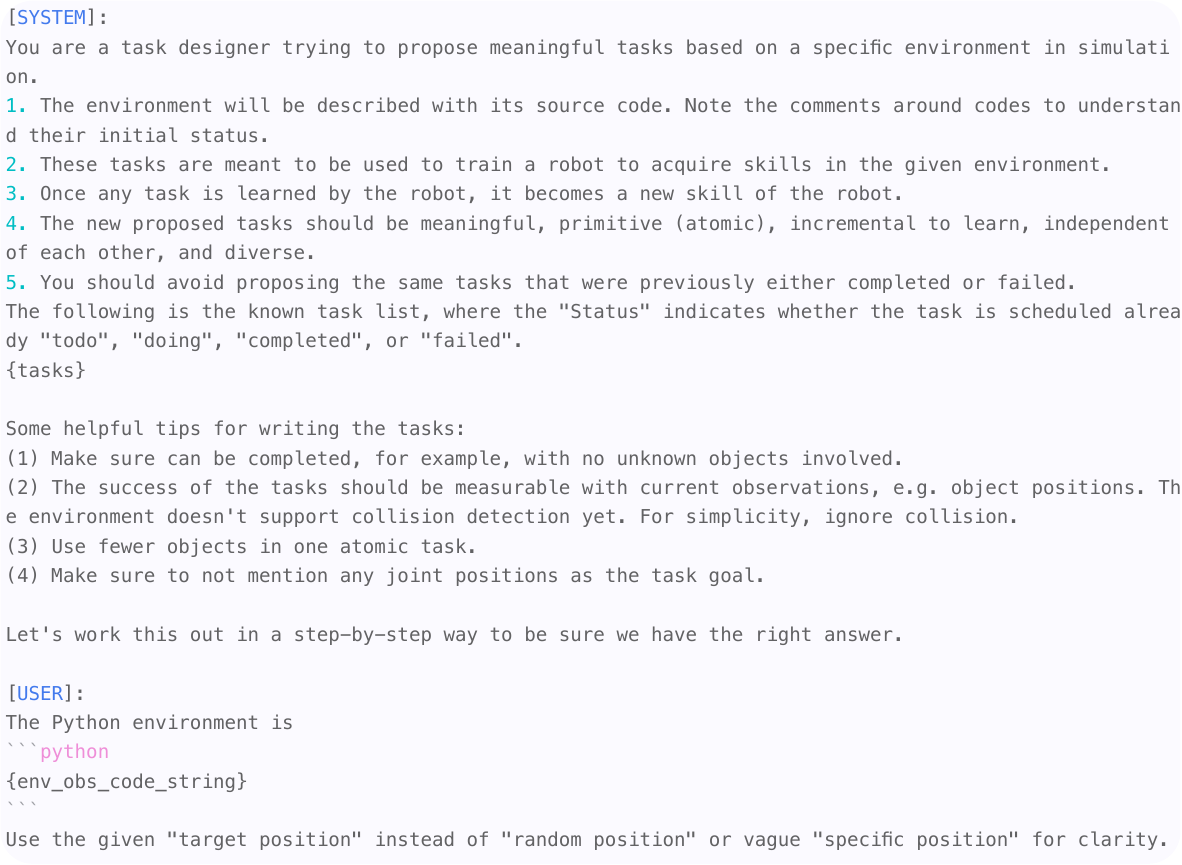}}
	\end{center}
    \caption{A snippet of task proposal prompt, where \texttt{\{tasks\}} points to previously proposed tasks, and \texttt{\{env\_obs\_code\_string\}} holds the place for incoming source codes for the environment.}
		\label{fig:prompt_proposal}
\end{figure}

See Fig.~\ref{fig:prompt_proposal}.

\subsection{Success Function and Reward Function Prompt}\label{app:prompt_succ}
\begin{figure}
	\begin{center}
		\centerline{\includegraphics[width=0.88\columnwidth]{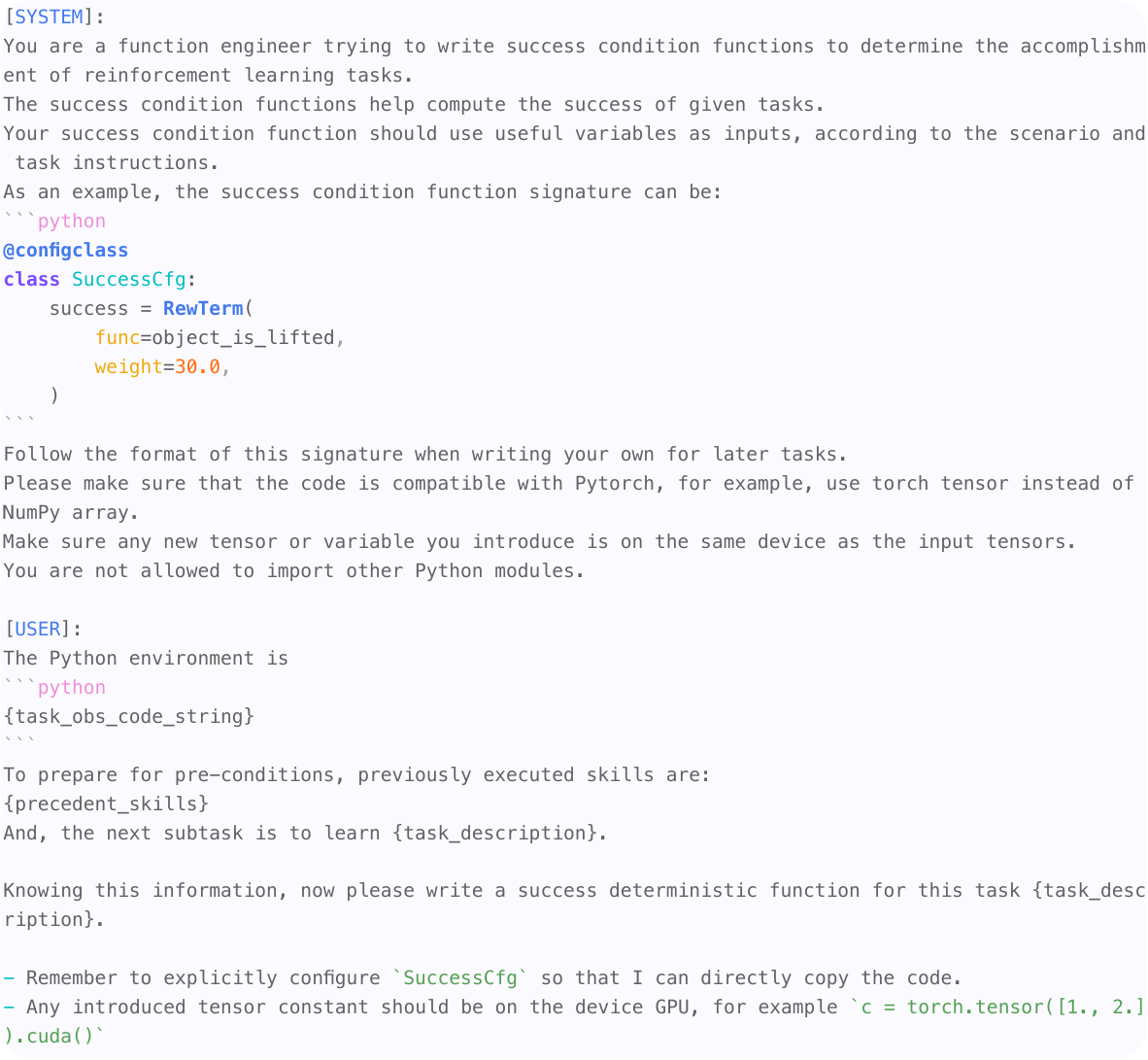}}
	\end{center}
  \caption{A prompt snippet for generating success functions given the environment information, where \texttt{\{precedent\_skills\}} holds the preceding executed skills as background information for the LLM to know the state that the learning will start with.}
		\label{fig:prompt_succ}
\end{figure}

\begin{figure}
	\begin{center}
		\centerline{\includegraphics[width=0.88\columnwidth]{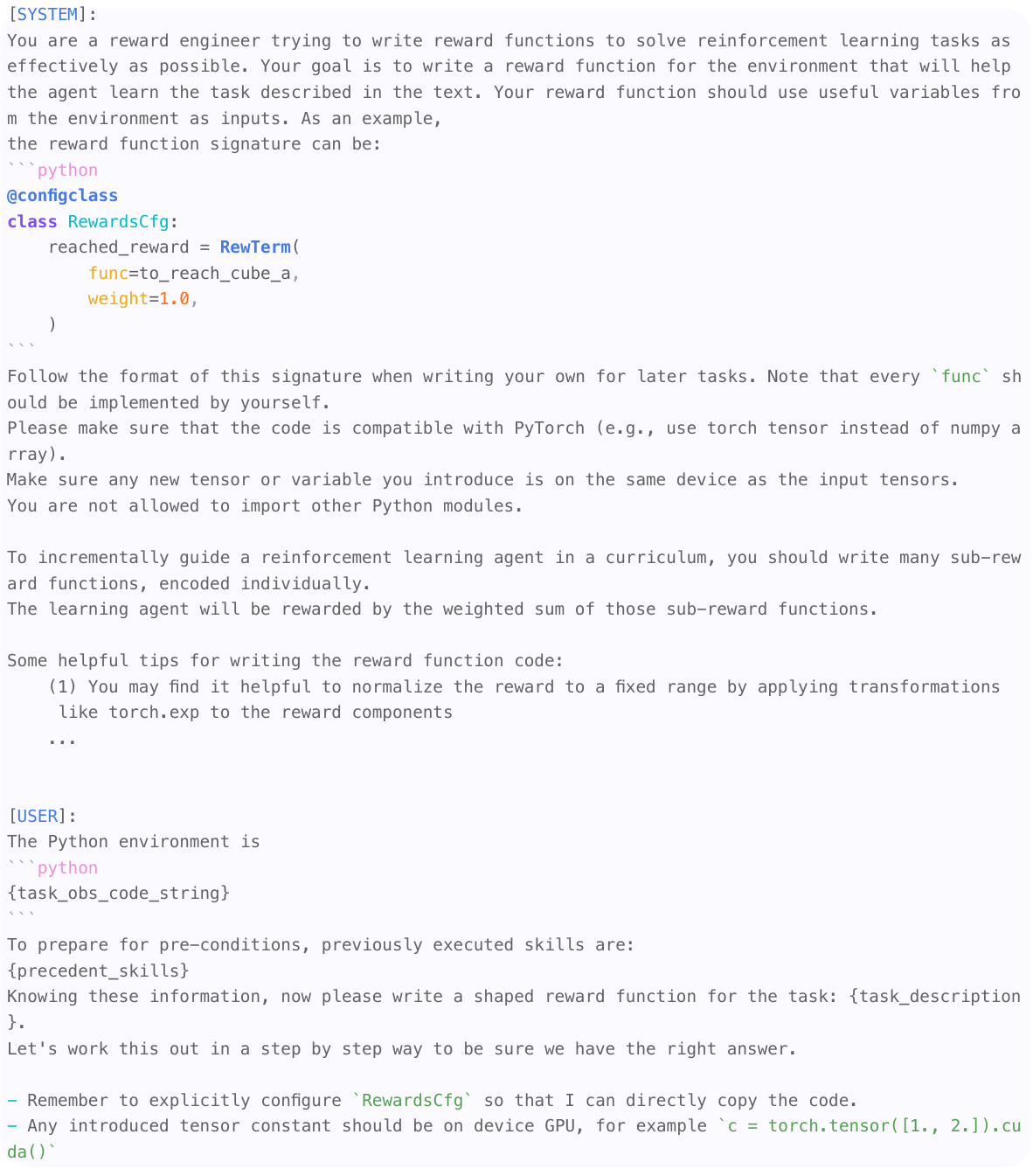}}
	\end{center}
    \caption{A prompt snippet for generating reward functions, which is similar to the prompt for the success function given the environment information. }
		\label{fig:prompt_reward}
\end{figure}

\begin{figure}
	\begin{center}
		\centerline{\includegraphics[width=0.88\columnwidth]{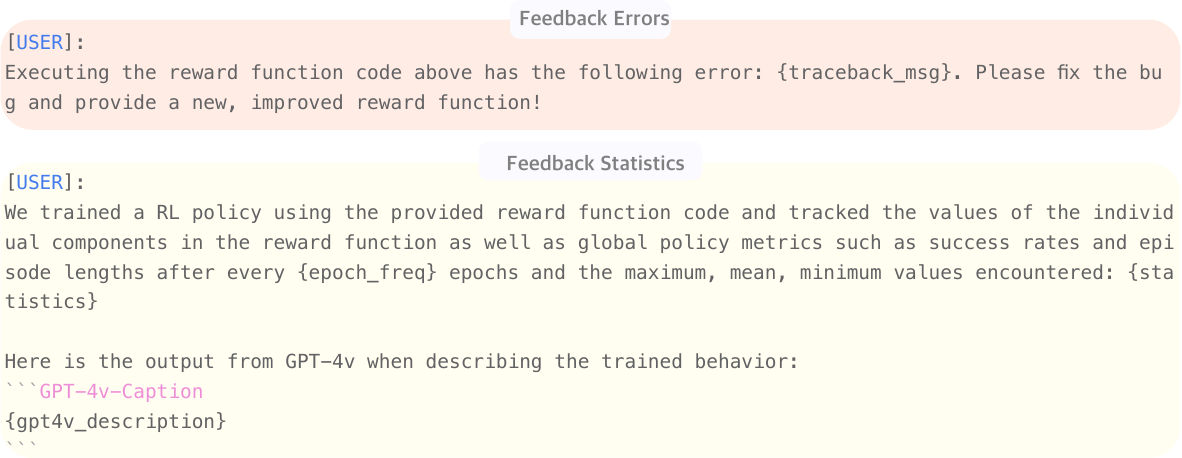}}
	\end{center}
    \caption{A prompt snippet for feeding back learning statistics and GPT-4V response for reward function iteration. \textit{Top}: if the code ended with an execution error, e.g., Pytorch tensor shape mismatch, the error messages will be fed back so the LLM can revise for a better one. \textit{Bottom}: if the code runs without bugs, the learning results will be collected for the iteration of reward functions, potentially resulting in more efficient ones.}
		\label{fig:prompt_reward_back}
\end{figure}

See Fig.~\ref{fig:prompt_succ}, Fig.~\ref{fig:prompt_reward} and Fig.\ref{fig:prompt_reward_back}.

\subsection{Behavior Assessment Prompt}\label{app:prompt_behavior}
\begin{figure}
	\begin{center}
		\centerline{\includegraphics[width=0.88\columnwidth]{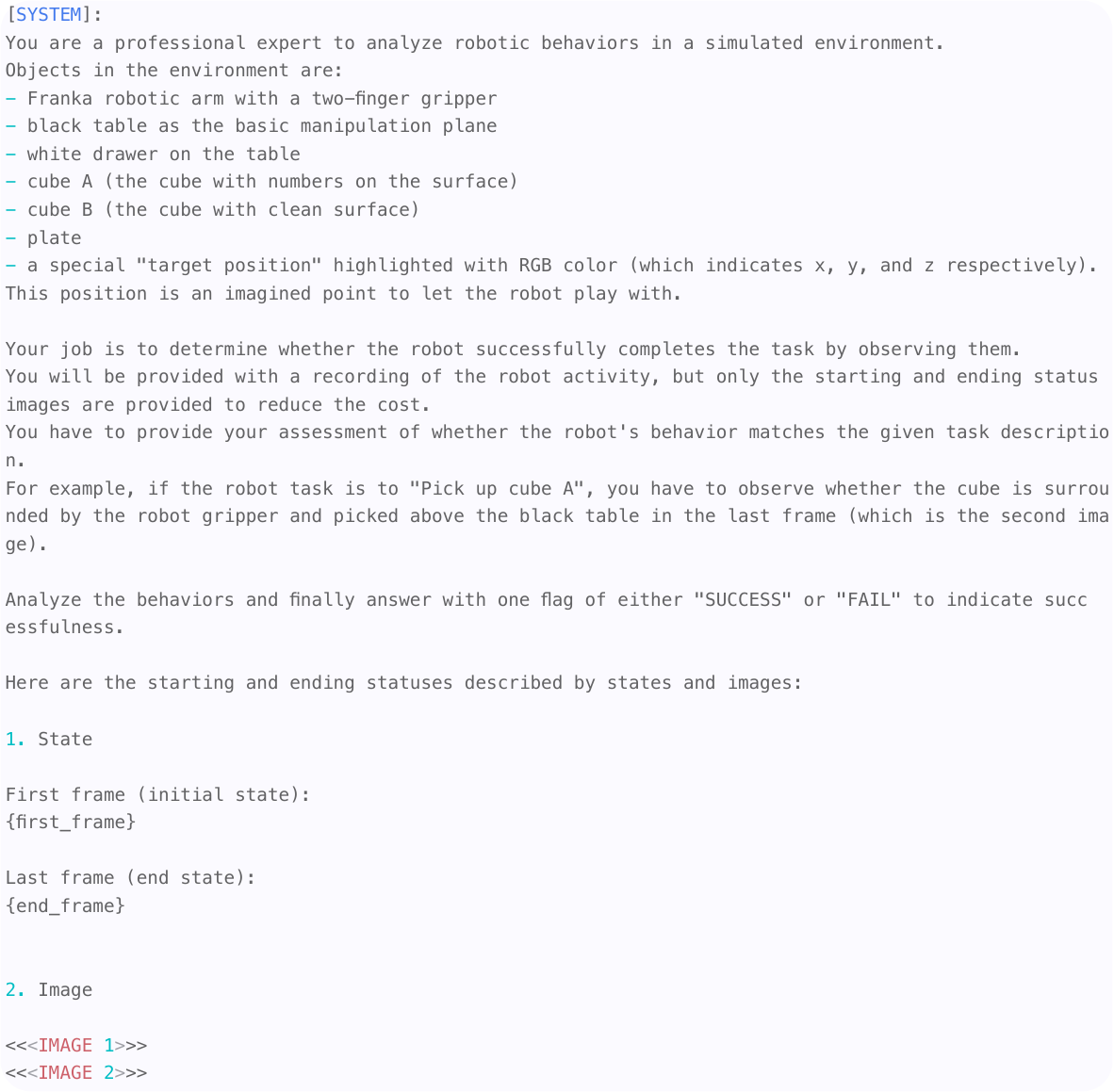}}
	\end{center}
  \caption{A snippet of prompts for robot behavior assessment using GPT-4V, where the \texttt{\{*\_frame\}} are state observations of the defined key frames of the recorded behavior video, and <<IMAGE X>> holds the place for corresponding key frame images.
    }
		\label{fig:prompt_behavior}
\end{figure}

See Fig.~\ref{fig:prompt_behavior}.

\end{document}